\documentclass[letterpaper]{article}
\usepackage{aaai2026}  % Remove 'submission' to enable author names
\usepackage{times}
\usepackage{helvet}
\usepackage{courier}
\usepackage[hyphens]{url}
\usepackage{graphicx}
\usepackage{natbib}
\usepackage{caption}
\usepackage{subcaption}
\usepackage{amsmath}
\usepackage{amssymb}
\usepackage{booktabs}
\usepackage{multirow}
\usepackage{inconsolata}
\usepackage{microtype}
\usepackage[utf8]{inputenc}
\usepackage{latexsym}
\usepackage{cleveref}
\usepackage{tablefootnote}

\usepackage{algorithm}
\usepackage{algorithmic}

\usepackage{newfloat}
\usepackage{listings}
\DeclareCaptionStyle{ruled}{labelfont=normalfont,labelsep=colon,strut=off}
\floatstyle{ruled}
\newfloat{listing}{tb}{lst}{}
\floatname{listing}{Listing}

\title{Mitigating Hallucination in Large Vision-Language Models via Adaptive Attention Calibration}

\author {
    Mehrdad Fazli,
    Bowen Wei,
    Ahmet Sari,
    Ziwei Zhu
}
\affiliations {
    Department of Computer Science, George Mason University, Fairfax, VA, USA\\
    \{mfazli, bwei2, asari2, zzhu20\}@gmu.edu
}

\begin{document}

\nocopyright

\maketitle

\begin{abstract}
Large vision-language models (LVLMs) achieve impressive performance on multimodal tasks but often suffer from hallucination, and confidently describe objects or attributes not present in the image. Current training-free interventions, struggle to maintain accuracy in open-ended and long-form generation scenarios. We introduce the Confidence-Aware Attention Calibration (CAAC) framework to address this challenge by targeting two key biases: spatial perception bias, which distributes attention disproportionately across image tokens, and modality bias, which shifts focus from visual to textual inputs over time. CAAC employs a two-step approach: Visual-Token Calibration (VTC) to balance attention across visual tokens, and Adaptive Attention Re-Scaling (AAR) to reinforce visual grounding guided by the model’s confidence. This confidence-driven adjustment ensures consistent visual alignment during generation. Experiments on CHAIR, AMBER, and POPE benchmarks demonstrate that CAAC outperforms baselines, particularly in long-form generations, effectively reducing hallucination. Data and code are available at~\url{https://github.com/mehrdadfazli/CAAC}.
\end{abstract}

% \section{some useful sentences}
% \begin{itemize}
%     \item We introduce an inherent bias existing in LVLMs that arises during the pretraining called "modality bias.
%     \item We define modality bias as the model's predisposition to favor one modality over another.
%     \item Through the lens of relevancy maps, we show that the model suffers from modality bias when generating long sequences of text.
%     \item Most of the efforts in LVLM hallucination is devoted to mitigating hallucination in discriminative tasks or short-response generative task while most of the hallucination actually happens in long generated sequences.
%     \item This selective intervention ensures AAR enhances visual grounding for image-dependent tokens without disrupting fluency for text-driven tokens.
% \end{itemize}

\section{Introduction}
Large vision-language models (LVLMs)~\cite{bai_qwen-vl_2023, chen_shikra_2023, liu_visual_2023, chen_internvl_2024, dai_instructblip_2023, ye_mplug-owl_2024} integrate visual and textual data using a pre-trained visual encoder, a cross-modal alignment module, and a powerful autoregressive decoder, enabling state-of-the-art performance in tasks such as image captioning, visual question answering, and visual reasoning. This multimodal capability has positioned LVLMs as key drivers in fields like content creation and human-computer interaction. However, a critical challenge is hallucination -- generating content ungrounded in the visual input, such as describing absent objects or misinterpreting scenes~\cite{bai_hallucination_2025,liu_survey_2024,li_evaluating_2023}. This undermines the reliability of LVLMs, posing significant barriers to their deployment in safety-critical domains like medical diagnosis and autonomous navigation.

\begin{figure}[t]
    \centering
    \includegraphics[width=\linewidth]{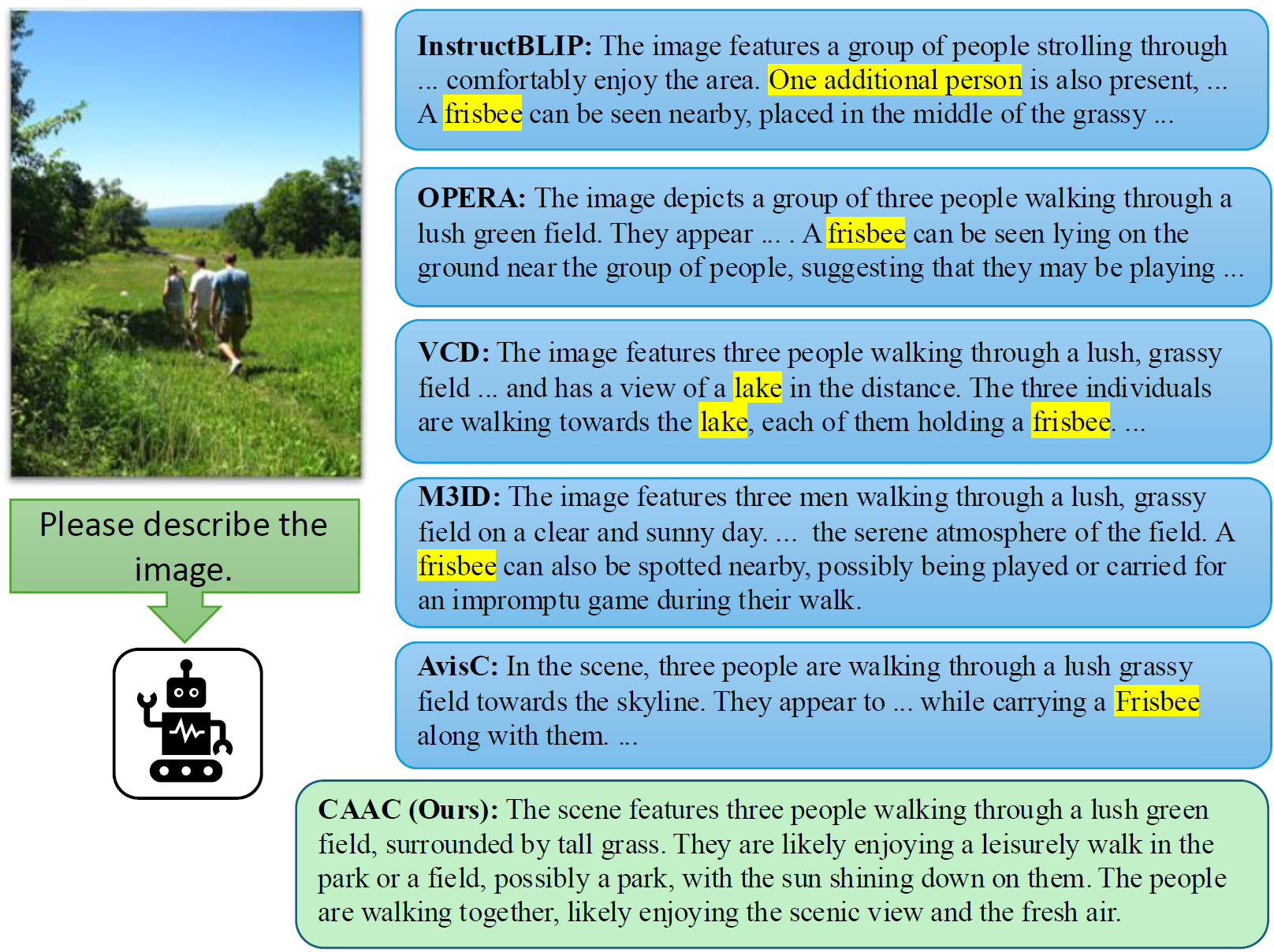}
    % \vspace{-10pt}
    \caption{Comparison of the long-form generation (Max Generated Tokens: 512) of the baseline methods and our proposed CAAC framework. Hallucinations are highlighted in yellow.}
    \label{fig:Case-Study}
    % \vspace{-20pt}
\end{figure}

Efforts to mitigate hallucination in LVLMs have spawned a rich body of research, with strategies broadly classified into three categories: fine-tuning~\cite{kim_exposing_2023, jiang_hallucination_2024, gunjal_detecting_2024}, post-hoc rectification~\cite{yin_woodpecker_2023, zhou_analyzing_2024}, and inference-time interventions~\cite{leng_mitigating_nodate, huang_opera_nodate}. Among them, inference-time interventions, due to their easy deployment and training-free nature, gained special momentum in the research community. Despite strong performance on discriminative tasks and short-form generation, existing methods struggle to maintain effectiveness in long-form generation. ~\Cref{fig:Case-Study} showcases an example of the failure of proposed hallucination mitigation methods under Max New Tokens of 512 (More qualitative examples are provided in Sec. 5 of the technical appendix). This limitation stems from two fundamental mechanisms of LVLMs. First, spatial perception bias results in disproportionate attention to specific image regions, causing the model to overlook relevant visual cues. Second, modality bias causes the model to increasingly allocate more attention to textual information over visual input as generation progresses, leading to content that is poorly grounded in the image. Both biases can significantly amplify the risk of hallucination in long-form generations.

To tackle these issues, we propose Confidence-Aware Attention Calibration (CAAC), a unified training-free approach to mitigate hallucinations by dynamically recalibrating the LVLM’s attention. CAAC uses the model’s token-level confidence to adjust the attention distribution adaptively. Specifically, it counteracts both spatial perception bias and modality bias in a two-step process: an initial calibration smooths the attention maps of the decoder to prevent over-concentration on any single image region, and a subsequent confidence-guided reweighting increases the influence of the visual input whenever the chance of hallucination is high. By continuously reinforcing visual information when the model is uncertain, CAAC preserves visual grounding throughout the generation. As a result, CAAC effectively curbs hallucinations, even in challenging open-ended and long-form generation tasks, without sacrificing the fluency or detail of the generated text.

Our main contributions are summarized: (1) \textbf{Hallucination Analysis:} We present a novel analysis of hallucination in LVLMs using relevancy maps, which reveals two root causes of ungrounded generation. (2) \textbf{Mitigation Method:} We propose CAAC, a training-free attention calibration framework, that adaptively calibrates the model’s attention to promote visual grounding. (3) \textbf{Performance Improvement:} We demonstrate that CAAC significantly reduces hallucinations on multiple benchmarks for open-ended image captioning. In particular, our method outperforms state-of-the-art baselines, achieving an average 4\% and 1.8\% reduction in the hallucination rate compared with the best baseline on the CHAIR and AMBER benchmarks, respectively. 
% Link to the Code and the data is provided in the technical appendix.
Code and data are available at~\url{https://github.com/mehrdadfazli/CAAC}.

\section{Related Work}
A more detailed discussion of the related works is provided in the technical appendix Sec. 3.

Large vision-language models (LVLMs) combine visual encoders like CLIP~\cite{radford_learning_2021} and ViT~\cite{fang_eva_2023}, cross-modal alignment modules such as linear projections~\cite{liu_visual_2023} or Q-formers~\cite{dai_instructblip_2023, zhu_minigpt-4_2023}, and language decoders like LLaMA~\cite{touvron_llama_2023} or Vicuna~\cite{zheng_judging_2023} to facilitate multimodal understanding. State-of-the-art models, including mPLUG-Owl2~\cite{ye_mplug-owl_2024}, InternVL~\cite{chen_internvl_2024}, and QwenVL~\cite{bai_qwen-vl_2023}, utilize optimized architectures and diverse datasets to achieve strong performance in tasks like image captioning and visual reasoning~\cite{xu_lvlm-ehub_2025}.

Hallucination in LVLMs occurs when generated outputs do not accurately reflect visual inputs, posing challenges to their reliability~\cite{guan_hallusionbench_2024, liu_survey_2024, bai_hallucination_2025}. Proposed mitigation strategies include fine-tuning techniques~\cite{kim_exposing_2023, jiang_hallucination_2024, liu_mitigating_2024, gunjal_detecting_2024}, post-hoc rectification methods~\cite{yin_woodpecker_2023, zhou_analyzing_2024}, and inference-time interventions~\cite{leng_mitigating_nodate, huang_opera_nodate, woo_dont_2024, suo_octopus_2025, favero_multi-modal_nodate}. Attention calibration, a training-free approach, has emerged as a promising solution to reduce hallucinations~\cite{zhu_mitigating_2025, zhang_seeing_2024, liu_paying_2024, gong_damro_2024, woo_dont_2024}. Our method builds on the insights derived from the previous works but introduces an adaptive intervention based on the model's confidence in predicting the next token.

\begin{figure}[t]
    \centering
    \includegraphics[width=0.6\linewidth]{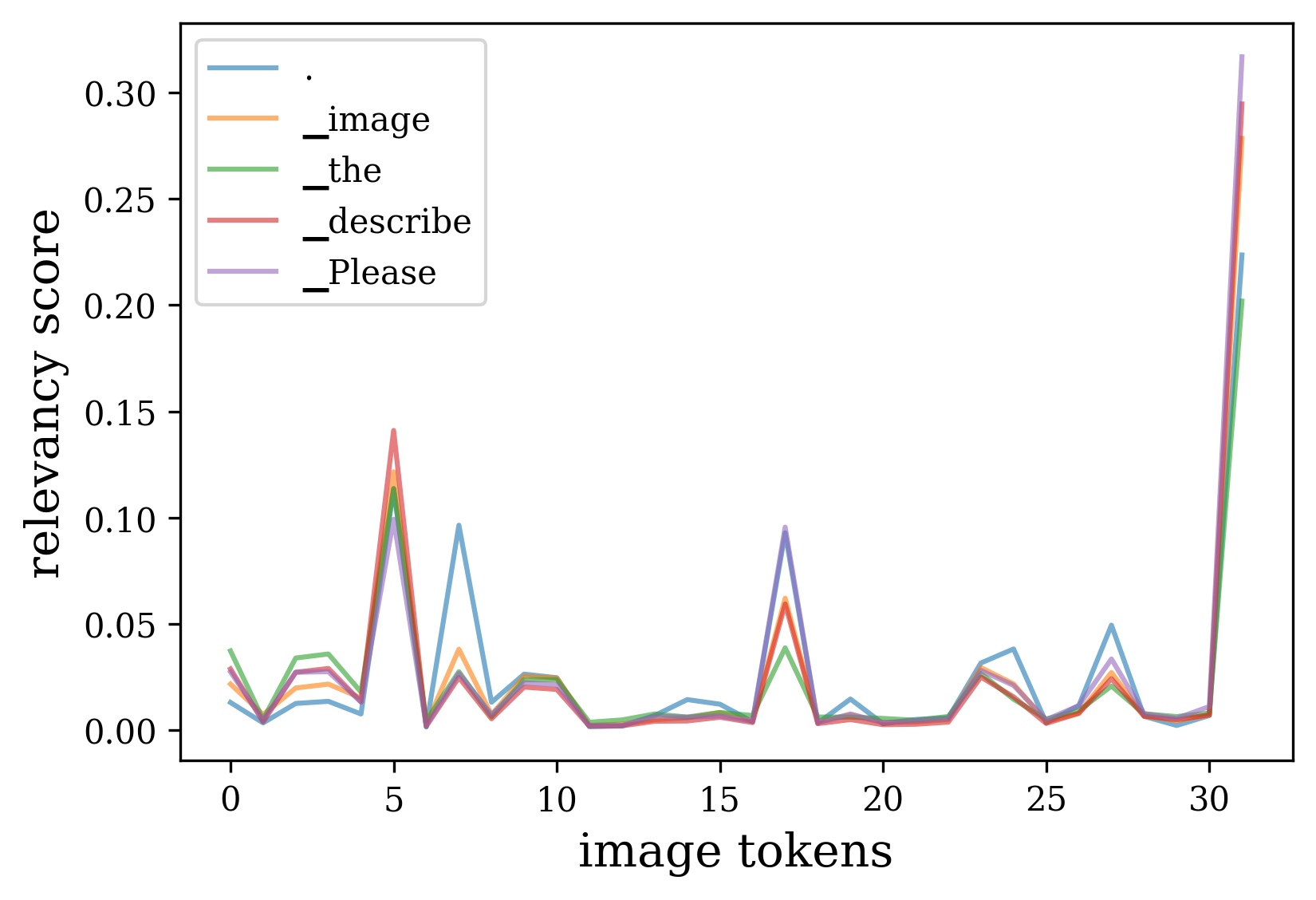}
    % % \vspace{-10pt}
    \caption{Distribution of image‑token relevancy scores for InstructBLIP given a black canvas as input image and the query "Please describe the image.". A pronounced skew toward a few image tokens can be witnessed.}
    \label{fig:instructblip-relevancy-skew}
    % % \vspace{-10pt}
\end{figure}

\begin{figure*}[t]
  \centering
  \begin{subfigure}[b]{0.32\textwidth}
    \centering
    \includegraphics[width=\linewidth]{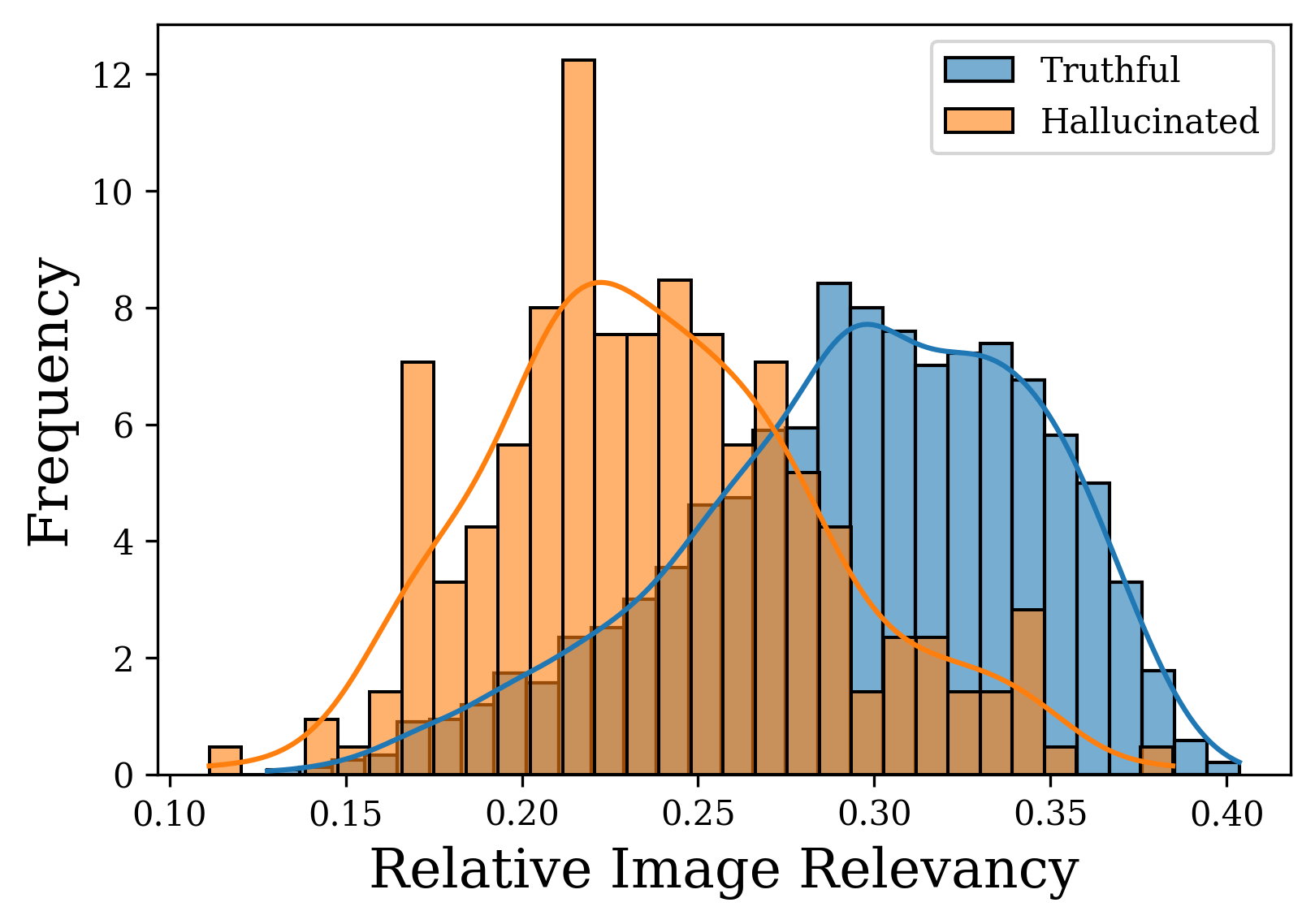}
    % % \vspace{-10pt}
    \caption{}
    \label{fig:rel-density}
  \end{subfigure}\hfill%
  \begin{subfigure}[b]{0.32\textwidth}
    \centering
    \includegraphics[width=\linewidth]{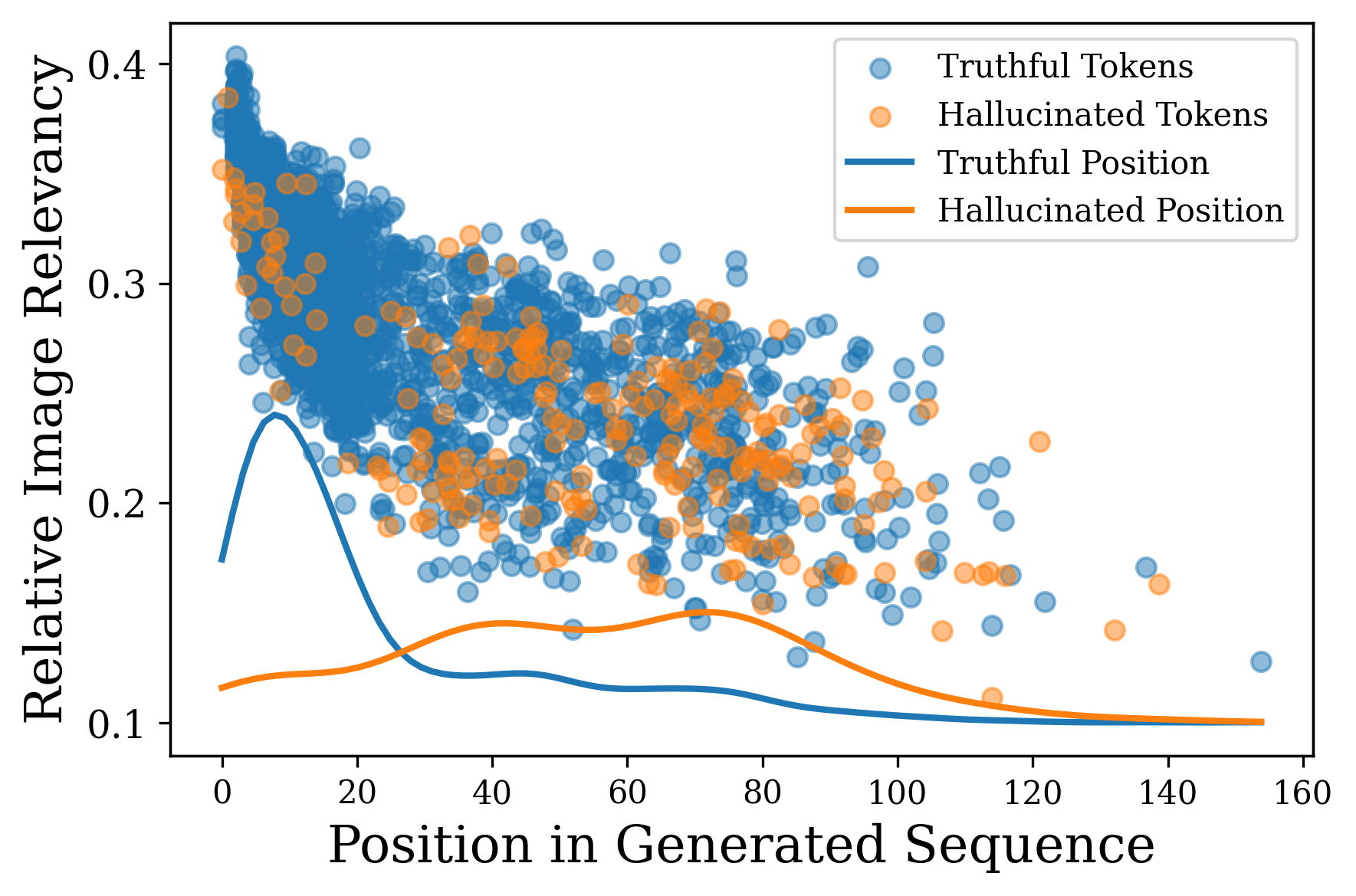}
    % % \vspace{-10pt}
    \caption{}
    \label{fig:rel-vs-pos}
  \end{subfigure}\hfill%
  \begin{subfigure}[b]{0.32\textwidth}
    \centering
    \includegraphics[width=\linewidth]{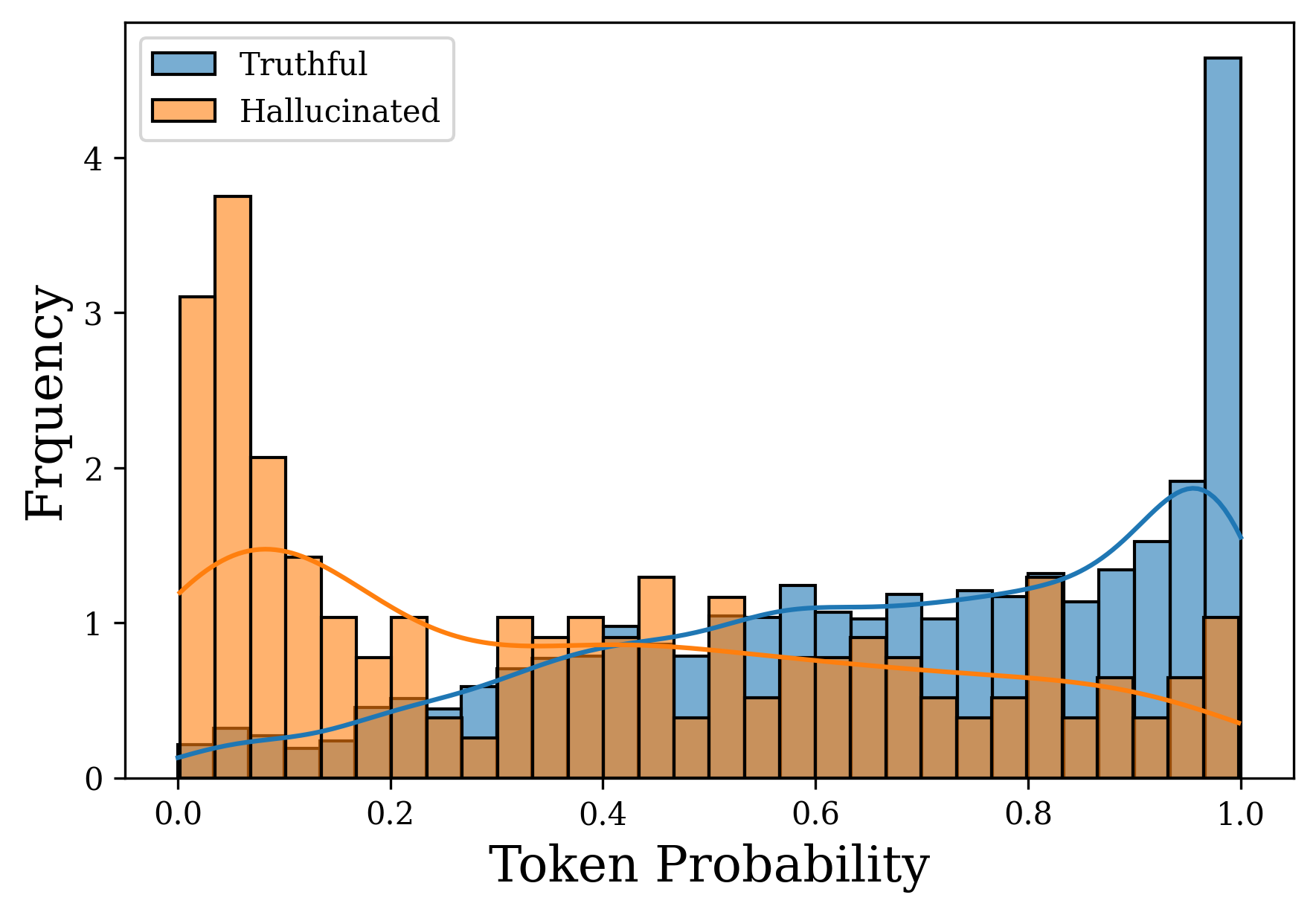}
    % % \vspace{-20pt}
    \caption{}
    \label{fig:prob-density}
  \end{subfigure}
  \caption{(a) Normalized histogram of relative image relevancy scores for truthful (blue) and hallucinatory (orange) tokens, showing higher image relevancy for truthful tokens. (b) Scatter plot of relative image relevancy vs. absolute position in the generated sequence. Every point represents one generated token (truthful or hallucinatory), and the lines indicate the density of token positions. (c) Normalized histogram of logit probabilities for truthful vs. hallucinatory tokens, showing lower probabilities for hallucinatory tokens. Best viewed in color.
    % (a) Distribution of relative image relevancy,  
    % (b) decay of mean image relevancy over generation,  
    % (c) distribution of logit probabilities for truthful vs.\ hallucinatory tokens.
  }
  % % \vspace{-10pt}
  \label{fig:decay-and-confidence}
\end{figure*}

\section{Proposed Method}

What causes LVLMs to describe objects or scenes absent from an image confidently? Our analysis identifies two primary culprits: spatial perception bias~\cite{zhu_mitigating_2025}, a skewed attention distribution favoring specific image tokens regardless of content, and modality bias, an increasing reliance on language priors over visual inputs as generation progresses. To tackle these challenges, we propose Confidence-Aware Attention Calibration (CAAC), which integrates two steps: an initial Visual-Token Calibration (VTC) to mitigate spatial perception bias by smoothing attention spikes across image tokens, and a confidence-driven Adaptive Attention Re-Scaling (AAR) to counteract modality bias by enhancing visual grounding throughout generation. %Next, we detail the inference process in LVLMs (\cref{inference-LVLM}), present our novel analysis using relevancy maps to uncover these biases (Sec. \ref{Analysis-1}, \ref{Analysis-2}), and introduce CAAC's components (\cref{CAAC-framework}), highlighting their synergistic design to improve reliability.

\subsection{Inference in LVLMs} \label{inference-LVLM}
Large vision–language models generate text conditioned on both an input image and a text prompt. An image is first encoded into visual tokens via a pre-trained vision encoder. The visual tokens are then mapped into the language embedding space using a linear projection or a more complex alignment module to extract textual information from the image, yielding image tokens \(I = \{i_1,\dots,i_{N_i}\}\). Concurrently, the text query is also tokenized into \(N_q\) tokens \(Q = \{q_1,\dots,q_{N_q}\}\). Then, the LLM decoder parameterized by $\theta$ receives concatenated embeddings \((I, Q)\) and auto‑regressively generates a sequence of \(N_g\) tokens \(G = \{y_1,\dots,y_{N_g}\}\). Formally, at $t$'th generation round, the next token is drawn from the following probability distribution:
\begin{equation}
    y_t \sim p_{\theta}(y_t|I,Q,y_{<t})
\end{equation}
where \(y_{<t} = \{y_1,\dots,y_{t-1}\}\) is the sequence of previously generated tokens. Various sampling strategies have been developed for efficient and controllable sampling from the probability distribution~\cite{shi_thorough_2024}. The generation process continues until the End-of-Sequence (EOS) token is selected or the maximum allowed number of tokens is reached.

\subsection{Analysis: Disproportionate attention across image tokens} \label{Analysis-1}

% Previous work has shown that LVLM decoders often concentrate attention on a small subset of visual tokens -- referred to as attention sinks\cite{zhang_seeing_2024}, summary tokens\cite{huang_opera_nodate}, or blind tokens\cite{woo_dont_2024} -- regardless of image content, even for blank inputs. This issue, also known as spatial perception bias\cite{zhu_mitigating_2025}, has been linked to downstream hallucinations~\cite{huang_opera_nodate, zhang_seeing_2024}. While our investigation shares a similar motivation with these studies, we identify a critical methodological flaw in their analysis: \textbf{they rely on raw attention weights from certain layers to quantify token importance}. However, in multi-head self-attention (MSA) architectures, token embeddings are deeply contextualized across layers, thus, the true token importance should be computed by propagating the contribution of each input token layer by layer based on attention distributions from each layer. To address this, we adopt \textbf{relevancy maps}~\cite{chefer_generic_2021}, which trace the generation of each output token back to the contribution of all input tokens, thereby offering a more faithful and holistic view of how the model leverages visual inputs. By using this more robust analytical tool, our work revisits and reevaluates prior claims, revealing new insights.

Previous studies have shown that LVLM decoders tend to concentrate attention on a small subset of visual tokens -- termed attention sinks~\cite{zhang_seeing_2024}, summary tokens~\cite{huang_opera_nodate}, or blind tokens~\cite{woo_dont_2024} -- regardless of image content, including blank inputs. This phenomenon, also known as spatial perception bias~\cite{zhu_mitigating_2025}, has been linked to downstream hallucination errors~\cite{huang_opera_nodate, zhang_seeing_2024}. While our analysis is motivated by similar concerns, we identify a key methodological limitation in prior work: \textbf{their conclusions are based on raw attention weights from individual layers}, which do not reliably reflect token importance. Indeed, token embeddings are progressively contextualized across layers, meaning that accurate attribution requires tracing the influence of each input token through the entire network.

To address this limitation, we leverage \textbf{relevancy maps}~\cite{chefer_generic_2021}, which propagate token-level contributions layer by layer, ultimately quantifying the influence of each input token on the generation of each output token. By adopting this more principled analysis, our work revisits and reinterprets previous findings, offering new insights. We observe that given a black canvas image and a standard query, less than 10\% of image tokens accumulate more than 50\% of relevancy scores, while the vast majority of image tokens contribute minimally (\Cref{fig:instructblip-relevancy-skew}). This distribution remains consistent across various meaningless inputs and queries (technical appendix Sec. 2), underscoring a robust bias pattern: \textbf{The decoder assigns disproportionate attention across image tokens, leading to the model’s over-reliance on a few image tokens, thereby increasing the likelihood of hallucination.}

% Previous work has demonstrated that LVLM decoders tend to funnel most attention into a small fraction of visual tokens, termed attention sinks~\cite{zhang_seeing_2024}, summary tokens~\cite{huang_opera_nodate}, or blind tokens~\cite{woo_dont_2024}, regardless of image content (even on blank images). This phenomenon is also known as the spatial perception bias~\cite{zhu_mitigating_2025}. Such tokens have been attributed to downstream hallucination errors in the literature~\cite{huang_opera_nodate, zhang_seeing_2024}. However, their analyses have a critical problem: \textbf{the analysis of disproportionate attention is conducted based on raw attention weights}. However, within the multi-head self-attention (MSA) module, token embeddings are contextualized across decoder layers, rendering single-layer attention weights an inadequate proxy for token contributions. 

% In contrast, we use relevancy maps~\cite{chefer_generic_2021}, which trace each generated token back to contributions of all input tokens, to obtain a more holistic view. 

\subsection{Analysis: Decaying attention to image tokens} \label{Analysis-2}

Another significant contributor to LVLM hallucination is the model’s increasing reliance on its text history at the expense of visual inputs, particularly in open-ended tasks like image captioning. Prior work has shown that when the model is uncertain, language priors often dominate the generation process~\cite{zhou_analyzing_2024}. To quantify this, we leverage AMBER’s generative pipeline, prompting InstructBLIP~\cite{dai_instructblip_2023} to describe each image in detail. Then, we extract truthful and hallucinatory tokens using predefined hallucinatory and truthful object sets from AMBER. We compute the \textit{relative image relevancy} by the relevancy map framework to quantify the aggregate contribution of all image tokens to the generation of each output token. For an input comprising \( I \) image tokens and \( T \) text tokens (total \( N = I + T \)), the relative image relevancy at generation step \( t \) is defined as:
\begin{equation}
    R_{rel_N} = {\sum_{i=1}^I R^{iN}}/{\sum_{j=1}^{N}R^{jN}}
\end{equation}
where $R^{ij}$ represents the influence of $i$'th token on $j$'th token. \Cref{fig:rel-density} shows the distribution of relative image relevancy for truthful and hallucinatory tokens. There is a statistically significant difference between the two distributions, suggesting that hallucinatory tokens have markedly lower relative image relevancy.
% We observe that hallucinatory tokens have markedly lower relative image relevancy compared to truthful tokens (Figure \ref{fig:rel-density}). 
Moreover, relative image relevancy declines as the generation lengthens (\cref{fig:rel-vs-pos}). This decay confirms that extended generation increases the model’s tendency to overlook visual inputs, a phenomenon we term modality bias, reflecting a preference for textual over visual information. The other takeaway is that the hallucinatory tokens appear later in the generated sequence, underscoring the importance of mitigating hallucinations in long-form generations.

We also examine the generation confidence by inspecting token logit probabilities (\cref{fig:prob-density}). We find that truthful tokens are heavily skewed toward high probabilities, whereas hallucinatory tokens are skewed toward the low-probability regime. It suggests a distinct generation dynamic between truthful and hallucinatory tokens: \textbf{the model hallucinates when its confidence is low and its attention to the image has diminished}.

\begin{figure*}[t]
    \centering
    \includegraphics[width=0.9\textwidth]{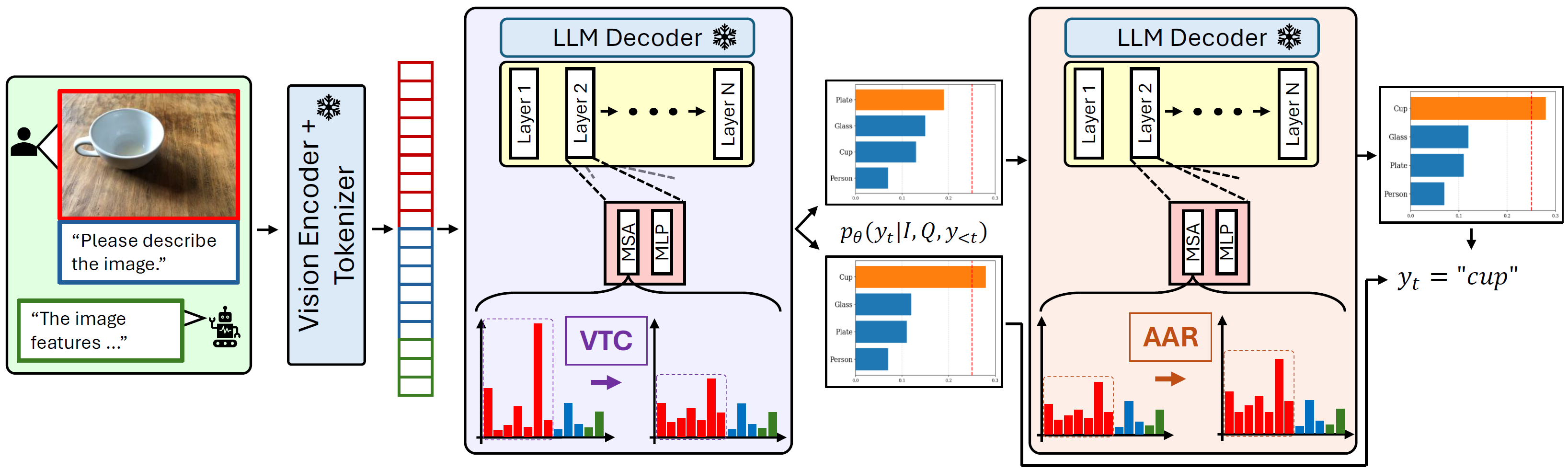}
    % % \vspace{-5pt}
    \caption{Overview of the CAAC Framework. The CAAC framework comprises two key components: VTC, which adjusts skewed attention to image tokens to reduce spatial perception bias, and AAR, which adaptively augments attention to image tokens to address modality bias. Both components are applied to the multi-head self-attention (MSA) module within the decoder.}
    % % \vspace{-10pt}
    \label{fig:enter-label}
\end{figure*}

\subsection{CAAC Framework} \label{CAAC-framework}
Our CAAC framework addresses two distinct biases operating in different dimensions within the LLM decoder. Spatial perception bias is a universal, query-agnostic distortion in attention distribution across image tokens. In contrast, modality bias operates at the token level, increasingly skewing attention toward textual inputs as generation length extends. CAAC tackles these challenges through a unified attention calibration strategy, featuring two components: Visual-Token Calibration (VTC), which corrects the universal spatial perception bias by adjusting attention weights, and Image Attention Upscaling (IAU), which mitigates modality bias by adaptively amplifying visual information during the generation. This integrated approach ensures a balanced multimodal processing, enhancing LVLM reliability.

\subsubsection{Visual-Token Calibration (VTC)}

VTC aims to mitigate spatial perception biases in LVLMs by adjusting the attention distribution over image tokens within the decoder's attention heads. By targeting the attention from the final query token to image tokens and applying a calibration derived from a reference input, we achieve a more balanced attention distribution while preserving essential visual information.
 
In LVLMs, the attention mechanism of the decoder plays a pivotal role in integrating visual and textual information. Specifically, the attention from the last query token to image tokens directly informs the prediction of the subsequent token, making it a critical point of intervention. Given an input comprising visual tokens \( I = \{i_1, i_2, \dots, i_{N_i}\} \) and query tokens \( Q = \{q_1, q_2, \dots, q_{N_q}\} \) ($N = I + Q$), the attention map for a given head \( h \) in layer \( l \) is denoted \( A^{h,l} \in \mathbb{R}^{(N_i + N_q) \times (N_i + N_q)} \). We focus on the submatrix corresponding to the last query token’s attention to image tokens, i.e., the last row’s first \( N_i \) columns, defined as \( V^{h,l} = [A^{h,l}_{N, j}]_{j\in \mathcal{I}} \in \mathbb{R}^{N_i} \).

\textbf{Calibration Vector Construction}:  
To establish a baseline for calibration, we use a reference input consisting of a meaningless image and a generic query (e.g., "What is this?"). Choosing a meaningless image ensures that attention patterns reflect the model’s baseline behavior rather than meaningful content, and empirical tests show that the choice of the meaningless image has no meaningful impact on the resulting calibration (technical appendix Sec. 2). For each attention head \( h \) in layer \( l \), we extract \( V^{h,l} \) from the reference input’s attention map. Alternatively, to enhance robustness, \( V^{h,l} \) may be computed as the average of the last few rows’ image-token columns. 

Therefore, given the vector \( V^{h,l} \in \mathbb{R}^{N_i} \), where \( V^{h,l} = [v_1, v_2, \ldots, v_{N_i}] \) and \( v_i \neq 0 \) for all \( i \), the initial inverse is computed as:
\begin{equation}
    V_{\text{cal},0}^{h,l} = [1/v_1, 1/v_2, \ldots, 1/v_{N_i}]
\end{equation}

To ensure the sum of entries remains consistent with the original vector, we scale \( V_{\text{cal},0}^{h,l} \) by the ratio of the sum of \( V^{h,l} \) to the sum of \( V_{\text{cal},0}^{h,l} \). The final calibration vector is thus:
\begin{equation}
    V_{\text{cal}}^{h,l} = \frac{\sum_{i=1}^{N_i} v_i}{\sum_{i=1}^{N_i} (1/v_i)} \cdot V_{\text{cal},0}^{h,l},
\end{equation}

where \( \sum_{i=1}^{N_i} v_i \) is the sum of the original attention weights, and \( \sum_{i=1}^{N_i} (1/v_i) \) is the sum of the initial inverted weights. Note that the product of \( V^{h,l} \) and \( V_{\text{cal}}^{h,l} \) results is a uniform vector with the same sum as \( V^{h,l} \). This inversion counteracts the skew attention pattern of the image tokens.

% The calibration vector \( V_{\text{cal}}^{h,l} \in \mathbb{R}^{N_i} \) is then derived by inverting \( V^{h,l} \) element-wise. Also, we make sure to preserve the sum of the entries in  \( V^{h,l} \):

% \begin{equation}
%     V_{\text{cal}}^{h,l} = \sum_{j=1}^I V^{h,l}_j  V^{h,l}^
% \end{equation}

\textbf{Application of Calibration}:  
For a specific input image and query pair, let \( V \in \mathbb{R}^{N_i} \) represent the attention from the last query token to image tokens in the attention map \( A^{h,l} \). We flatten this by computing the element-wise product \( V_u = V \odot V_{\text{cal}}^{h,l} \), where \( \odot \) denotes the Hadamard product. \( V_u \) approximates a uniform attention distribution across image tokens. However, enforcing strict uniformity can distort visual information, as positional embeddings naturally differentiate image token representations, even for identical patches. This differentiation is naturally reflected in the attention scores received by different image tokens.

\textbf{Smoothing with Parameter \( \beta \)}:  
To balance bias correction and information preservation, we introduce a smoothing parameter \( \beta \in [0,1] \) to control smoothing. The smoothed attention vector \( V_s \) is computed as a weighted average of the original and calibrated vectors:
\begin{equation}
    V_s = (1 - \beta) V + \beta V_u
\end{equation}

When \( \beta = 0 \), the original attention \( V \) is retained and when \( \beta = 1 \), the fully calibrated \( V_u \) is applied, yielding a near-uniform distribution. Intermediate values of \( \beta \) allow for promoting more balanced attention distribution without over-correcting the attention distribution. This flexibility ensures that the calibration enhances model reliability and is what makes the VTC module different than UAC~\cite{zhu_mitigating_2025}.

\subsubsection{Adaptive Attention Re-Scaling (AAR)}

AAR is designed to mitigate modality bias, where attention to image tokens diminishes over time during autoregressive generation. AAR counteracts this by dynamically increasing the attention from the last query token to image tokens, reinforcing visual grounding throughout the generation sequence, particularly when the model’s predictions falter.  
AAR focuses on the same segment of the attention map as the VTC module, specifically the attention vector \( V^{h,l} = [A^{h,l}_{N, j}]_{j\in \mathcal{I}} \in \mathbb{R}^{N_i} \) to steer model's attention toward visual information by scaling up the attention weights of visual tokens.

% Explaining the confidence-driven scaling mechanism

\textbf{Confidence-Aware Scaling}:  
AAR operates autoregressively, adjusting attention in every generation round to maintain visual relevance across the entire sequence. A key question is: \textit{what is the appropriate scaling factor, as token dependency on visual input varies?} Tokens essential for text cohesion (e.g., conjunctions) require minimal intervention, whereas image-dependent tokens (e.g., nouns and adjectives describing visual content) demand stronger visual grounding. Our analysis revealed that hallucinatory tokens often emerge when the model lacks confidence (\cref{fig:prob-density}). This insight drives AAR’s adaptive strategy: \textbf{scaling is triggered by the model’s uncertainty}.

In generation round \(t\), a forward pass computes the maximum logit probability \( p_t \) for the predicted token.
\begin{equation}
    p_t \;=\; \max_{y}\;p_{\theta}\bigl(y\mid I,Q,y_{<t}\bigr).
\end{equation}

If \( p_t \) falls below a preset threshold \( p_{\text{thr}} \), AAR calculates a scaling factor \( \lambda \) as a probability-weighted average of set minimum and maximum scale factor:
\begin{equation}
    \lambda_t = \lambda_{\text{min}} \cdot p + \lambda_{\text{max}} \cdot (1 - p)
\end{equation}

With \( \lambda_{\text{min}} = 1 \) we ensure no scaling is applied when the model is fully confident (\( p = 1 \)), while \( \lambda_{\text{max}} \) sets the upper bound for scaling when confidence is minimal (\( p = 0 \)). As \( p \) decreases, \( \lambda \) increases, amplifying attention to image tokens precisely when hallucination risk is highest.

% Detailing the application of AAR
\textbf{Application of AAR}:  
As AAR is bound to change the sum of the row it is applied to, we need to apply it to the attention weights before softmax. After the intervention, softmax is applied to ensure all rows sum to 1. When \( p < p_{\text{thr}} \), the attention vector before softmax \( V^{h,l} \) is scaled:
\begin{equation}
    V_{\text{t, scaled}}^{h,l} = \lambda_t \cdot V_t^{h,l}
\end{equation}

This scaled vector replaces the original vector in the decoder’s attention mechanism, shifting focus toward visual inputs. If \( p \geq p_{\text{thr}} \), no scaling occurs, preserving the model’s natural behavior.

% for high-confidence predictions. This selective intervention ensures AAR enhances visual grounding for image-dependent tokens without disrupting fluency for text-driven tokens.

\begin{table}[t]\small
  \centering
  \caption{Performance on CHAIR Benchmark}
  % % \vspace{-5pt}
  \begin{tabular}{lcc|cc|cc}
    \toprule
    \multirow{2}{*}{Method} & \multicolumn{2}{c|}{LLaVA-1.5} & \multicolumn{2}{c|}{InstructBLIP} & \multicolumn{2}{c}{LLaVA-NeXT} \\
    % \cline{2-7}
     & \(\mathrm{C}_s\)$\downarrow$ & \(\mathrm{C}_i\)$\downarrow$ & \(\mathrm{C}_s\)$\downarrow$ & \(\mathrm{C}_i\)$\downarrow$ & \(\mathrm{C}_s\)$\downarrow$ & \(\mathrm{C}_i\)$\downarrow$ \\
    \midrule
    base model & 55.2 & 17.6 & 55.6 & 16.6 & 33 & 9.4 \\
    + OPERA     & \underline{44.6} & \underline{12.8} & \underline{46.4} & \underline{14.2} & -- & -- \\
    + VCD       & 57.8 & 16.3 & 60.8 & 17.9 & 41.6 & 9.9 \\
    + AvisC     & 60.4 & 17.2 & 71.0 & 20.1 & 34.8 & 9.3 \\
    + M3ID      & 56.2 & 16.4 & 72.8 & 21.1 & 42 & 12.4 \\
    + CAAC      & \textbf{39.2} & \textbf{10.4} & \textbf{37.4} & \textbf{10.8} & \textbf{30.6} & \textbf{8.1} \\
    \bottomrule
  \end{tabular}
  \label{tab:chair_results}
  % % \vspace{-5pt}
\end{table}

\section{Experimental Results}
\subsection{Setup} \label{exp-setup}
% This section outlines the experimental setup used to evaluate the CAAC framework, detailing the selected models, benchmarks, baselines, and implementation specifics.
\noindent\textbf{Models.}
We evaluate CAAC on three 7B-parameter LVLMs: InstructBLIP, LLaVA-1.5, and LLaVA-NeXT, selected for direct comparison with baselines~\cite{leng_mitigating_nodate,huang_opera_nodate, favero_multi-modal_nodate}. However, our CAAC framework is model-agnostic and can be seamlessly integrated with any LVLM. Experimental settings and implementation details are presented in the technical appendix (Sec. 1).

\noindent\textbf{Benchmarks.} 
We prioritize generative benchmarks that support open-ended generations. We adopt CHAIR \citep{rohrbach_object_2019} and AMBER \citep{wang_amber_2024} as our generative benchmarks, and POPE MSCOCO \citep{li_evaluating_2023} as the discriminative benchmark to provide a comprehensive evaluation of CAAC.

\noindent\textbf{Metrics.}
We mainly focus on metrics that directly measure hallucination rates, such as \(\mathrm{CHAIR}_i\) and \(\mathrm{CHAIR}_s\) for the CHAIR benchmark, and \(\mathrm{CHAIR}\) and \(\mathrm{HAL}\) for the AMBER benchmark, due to their critical role in assessing the model’s factual alignment with visual input. We also report \(\mathrm{COVER}\) scores for AMBER, which measures the informativeness and completeness of generated responses, and accuracy and F-1 score for the POPE benchmark. However, we note that high \(\mathrm{COVER}\) scores paired with elevated hallucination rates are undesired in many real-world applications~\cite{keskar_evaluating_nodate, magesh_hallucination-free_2025, hartsock_vision-language_2024}, as the model may generate exhaustive but factually incorrect descriptions. The goal is to \textit{maximally reduce hallucination metrics while maintaining high coverage values}. 
% For the POPE benchmark, we report Accuracy and \(\mathrm{F1}\) scores to complement our evaluation.

\noindent\textbf{Baselines.}  
Baselines include three training-free contrastive decoding methods, VCD~\cite{leng_mitigating_nodate}, AvisC~\cite{woo_dont_2024}, and M3ID~\cite{favero_multi-modal_nodate}, which leverage the contrastive decoding technique~\cite{li_contrastive_2023}, and OPERA~\cite{huang_opera_nodate}, a beam‑search modification that penalizes over‑trusted tokens to promote visual grounding. We were unable to reproduce OPERA’s results on LLaVA-NeXT due to compatibility challenges in adapting its inference-time beam search to the updated Hugging Face generation API, compounded by LLaVA-NeXT’s use of a dynamic number of image tokens.

% \noindent\textbf{Implementation Details.}  
% For the baselines, we adopt the hyperparameter settings reported in their respective papers to ensure consistency. For CAAC, we set the smoothing parameter \( \beta \) to 0.7 for LLaVA, 0.3 for LLaVA-NeXT, and 0.5 for InstructBLIP. The maximum scaling factor for AAR is set to \( \lambda_{\text{max}} = 1.5 \) for both tasks, with \( \lambda_{\text{min}} = 1.0 \) and \(p_{thr}=0.25 \). More experimental details have been presented in the supplementary material (Sec. 1).

\subsection{Comparison to Baselines}

\begin{table*}[t]\small
  \centering
  \caption{Performance on AMBER Benchmark Across Different MaxTokens Settings}
  
  % \caption*{\footnotesize * OPERA results for LLaVA-NeXT are omitted due to reproducibility issues.}
  % % \vspace{-5pt}
  \begin{tabular}{l|ccc|ccc|ccc}
    \toprule
    \multirow{2}{*}{Mitigation Method} & \multicolumn{3}{c|}{MaxTokens 64} & \multicolumn{3}{c|}{MaxTokens 512} & \multicolumn{3}{c}{AVG}\\
     & CHAIR$\downarrow$ & HAL$\downarrow$ & COVER$\uparrow$ & CHAIR$\downarrow$ & HAL$\downarrow$ & COVER$\uparrow$ & CHAIR$\downarrow$ & HAL$\downarrow$ & COVER$\uparrow$\\
    
    \midrule
    InstructBLIP    &  9.6  & 36.0  & 46.5 & 12.8  & 53.5 & 52.7  & 11.2 & 44.8 & 49.6\\
    + OPERA         &  \underline{6.6}  & \underline{24.7}  & 46.4 & \underline{9.7}  & \underline{40.5} & 51.2 & \underline{8.2} & \underline{32.6} & 48.8\\
    + VCD           &  7.6  & 29.9  & \underline{47.5} & 10.8  & 46.6  & \textbf{53.4} & 9.2 & 38.3 & \textbf{50.5}\\
    + M3ID          &  6.9 & 27.5  & 47.2 &  10.4 &  47.3 &  51.7 & 8.7 & 37.4 & 49.5\\
    % + Octopus       &  \underline{6.10} & 48.50 & 22.2 & 1.3 &  --   &  --   &  -- & - \\
    + AvisC         &  6.7  & 28.0 & 46.7 & 10.1  & 46.8 & 51.2 & 8.4 & 37.4 & 49.0\\
    % + CAAC (Ours)   & \textbf{4.8} & \textbf{20.3}  & 46 & \textbf{5.6}  & \textbf{25.8}  & 47.8 & \textbf{5.2} & \textbf{23.1} & 46.9\\
    + CAAC   & \textbf{5.2} & \textbf{20.5} & \textbf{48.2} & \textbf{7.0}  & \textbf{30.9} & \underline{51.9} & \textbf{6.1} & \textbf{25.7} & \underline{50.1} \\
    \midrule
    LLaVA-1.5         &  8.0  & 31.0  & 44.5 & 11.3  & 48.1  & 50.4 & 9.6 & 39.5 & 47.5 \\
    + OPERA       &  \underline{5.1} & \textbf{19.1}  & 45.0 &  7.3  & \underline{29.5}  & 47.5 & \underline{6.2} & \underline{24.3} & 46.3 \\
    + VCD         &  6.7 & 27.8  & \underline{46.5} & 8.2  & 37.3  & 51.9 & 7.5 & 32.5 & 49.2\\
    + M3ID        &  6.0 & 26.0   & \textbf{48.9} &  \underline{7.2}   &  41.4 &  \textbf{57.3} & 6.6 & 33.7 & \textbf{53.1}\\
    % + Octopus     &  \textbf{4.80} & 49.20 & 23.4 &  1.2   &  -   &  - & - \\
    + AvisC       &  6.3 & 25.6 & \underline{46.5} & 11.0  & 48.0 & \underline{52.5} & 8.6 & 36.8 & \underline{49.5}\\
    % + CAAC (Ours) & \textbf{4.90} & \underline{19.7}  & 45.40 & \textbf{6.0} & \textbf{24.8} & 47.6 & \textbf{5.5} & \textbf{22.3} & 46.5\\
    + CAAC & \textbf{5.0} & \underline{20.1}  & \underline{46.5} & \textbf{6.0} & \textbf{25.0} & 48.7 & \textbf{5.5} & \textbf{22.6} & 47.6\\
    \midrule
    LLaVA-NeXT    &  6.5  & \underline{20.6}  & 35.5 & 9.3  & 51.3  & 60.6 & 7.9 & 36.0 & 48.1 \\
    + OPERA       &  -  & -  & - & -  & -  & - & - & - & - \\
    + VCD         &  8.0  & 26.4  & \textbf{38.4} & 10.5  & 57.2  & \textbf{63.5} & 9.3 & 41.8 & \textbf{51.0} \\
    + M3ID        &  7.5  & 23.2  & \underline{37.8} & 12.4  & 59.8  & \underline{61.4} & 10.0 & 41.5 & \underline{49.6} \\
    + AvisC       &  \underline{6.3}  & \textbf{19.9}  & 36 & \underline{9.2}  & \underline{50.4}  & 61.1 & \underline{7.8} & \underline{35.2} & 48.6 \\
    + CAAC        &  \textbf{6.0}  & \textbf{19.9}  & 37.3 & \textbf{8.8}  & \textbf{47.5}  & 60.5 & \textbf{7.4} & \textbf{33.7} & 48.9 \\

    \bottomrule
  \end{tabular}
  % % \vspace{-10pt}
  \label{tab:amber_results}
\end{table*}

\begin{table*}[t]\small
  \centering
  \caption{Performance on POPE MSCOCO Benchmark Across Different Sampling Settings}
  % % \vspace{-5pt}
  \begin{tabular}{lcccccccc}
    \toprule
    \multirow{2}{*}{Mitigation Method} & \multicolumn{2}{c}{Random} & \multicolumn{2}{c}{Popular} & \multicolumn{2}{c}{Adversarial} & \multicolumn{2}{c}{AVG} \\
    & Accuracy & F1 & Accuracy & F1 & Accuracy & F1 & Accuracy & F1 \\
    \midrule
    InstructBLIP & 81.5 & 81.2 & 78.5 & 78.8 & 77.4 & 78.0 & 79.1 & 79.3 \\
    + OPERA & \textbf{89.2} & \textbf{88.7} & \underline{84.0} & \textbf{83.7} & \textbf{81.8} & \textbf{81.9} & \textbf{85.0} & \textbf{84.8} \\
    + VCD & 82.0 & 81.6 & 79.1 & 79.2 & 77.2 & 77.7 & 79.4 & 79.5 \\
    + M3ID & 82.3 & 81.5 & 80.9 & 80.4 & 78.5 & 78.5 & 80.6 & 80.1 \\
    % + Octopus & 86.6 & 85.3 & \textbf{84.9} & \textbf{83.6} & \textbf{82.8} & 81.4 & 84.8 & 83.4 \\
    + AvisC & 86.0 & 84.4 & \textbf{84.3} & 82.8 & \textbf{81.8} & 80.7 & 84.0 & 82.6 \\
    + CAAC (Ours) & \underline{87.7} & \underline{87.1} & 83.5 & \underline{83.4} & \underline{81.2} & \underline{81.5} & \underline{84.1} & \underline{84.0} \\
    \midrule
    LLaVA-1.5 & 83.8 & 81.9 & 82.6 & 80.9 & 79.8 & 78.5 & 82.1 & 80.4 \\
    + OPERA & \textbf{88.5} & \textbf{88.5} & \underline{85.6} & \textbf{85.6} & 80.8 & \textbf{81.7} & \underline{85.0} & \textbf{85.3} \\
    + VCD & 85.4 & 84.0 & 83.2 & 81.9 & 80.3 & 79.5 & 83.0 & 81.8 \\
    + M3ID & 86.1 & 81.9 & 82.1 & 80.8 & 79.5 & 78.2 & 82.6 & 80.3 \\
    % + Octopus & 87.5 & 85.4 & 83.2 & 84.2 & \textbf{82.2} & \textbf{81.4} & 84.3 & 83.7 \\
    + AvisC & 84.7 & 82.2 & 83.7 & 81.3 & \textbf{81.8} & 79.6 & 83.4 & 81.0 \\
    + CAAC (Ours) & \textbf{88.5} & \underline{87.8} & \textbf{85.9} & \underline{85.5} & \underline{81.0} & \underline{81.4} & \textbf{85.1} & \underline{84.9} \\
    \midrule
    LLaVA-NeXT & 84.5 & 85.8 & 86.5 & 84.9 & \underline{85.6} & 84 & 86.5 & 84.9 \\
    + OPERA & - & - & - & - & - & - & - & -  \\
    + VCD & \textbf{88.1} & \textbf{86.8} & \textbf{87.0} & \textbf{85.9} & 85.0 & \underline{84.0} & \underline{86.7} & \textbf{85.6}  \\
    + M3ID & 85.3 & 82.8 & 84.7 & 82.3 & 84.0 & 81.7 & 84.7 & 82.3  \\
    + AvisC & 87.3 & 85.6 & 86.5 & 84.8 & 85.3 & 83.9 & 86.4 & 84.8  \\
    + CAAC (Ours) & \underline{87.7} & \underline{86.2} & \underline{86.8} & \underline{85.3} & \textbf{85.8} & \textbf{84.4} & \textbf{86.8} & \underline{85.3}  \\
    \bottomrule
    % % \vspace{-10pt}
  \end{tabular}
  \label{tab:pope_results}
\end{table*}

% We used the same evaluation setting as OPERA. We also used the same subset of 500 images from the validation set of the COCO 2014 dataset~\cite{lin_microsoft_2015}, paired them with the prompt "Please describe this image in detail.", and collected the responses from LVLM. 
\paragraph{CHAIR.}
The CHAIR benchmark~\cite{rohrbach_object_2019} evaluates object hallucination in image captioning by calculating two metrics on MSCOCO 2014 images~\cite{lin_microsoft_2015}: \(\mathrm{CHAIR}_i\), the proportion of hallucinated objects relative to all mentioned objects, and \(\mathrm{CHAIR}_s\), the ratio of sentences that containing hallucination. We set Max Tokens to 512 to avoid prematurely truncating generation sequences. \cref{tab:chair_results} summarizes the results of the CAAC framework and the baselines on the CHAIR benchmark. As shown, CAAC effectively reduces the hallucination rates, \(\mathrm{CHAIR}_i\) and \(\mathrm{CHAIR}_s\), compared to the baselines. 
% We note that OPERA results are not provided for LLaVA-NeXT~\cite{huang_opera_nodate}.

\paragraph{AMBER.} 
The AMBER benchmark~\citep{wang_amber_2024} assesses hallucinations in LVLMs through generative and discriminative tasks, focusing on object existence, attributes, and relationships. We focus on the generative task, conducting experiments under Max Tokens 64, aligning with baseline configurations, and Max Tokens 512 for longer generations. AMBER uses three metrics: CHAIR (frequency of hallucinated objects), HAL (proportion of responses containing hallucinations), and COVER (proportion of image objects mentioned) to evaluate faithfulness and completeness.

Our CAAC framework excels on the AMBER benchmark, delivering the lowest hallucination rates in CHAIR and HAL metrics across all settings and models (\cref{tab:amber_results}). \textit{Contrastive decoding techniques, however, show significant degradation in managing hallucinations during long generations (MaxTokens 512)}, underscoring their limitations. While CAAC does not deliver the best COVER in some settings, it maintains a high level of COVER, better than the base model. CD methods' high COVER scores come at the cost of more hallucinations. For example, M3ID has the highest COVER score with InstructBLIP, but it also has the highest CHAIR and HAL scores. Notably, CAAC outperforms OPERA, the best-performing baseline in terms of hallucination rate, in the COVER scores.

% While CAAC lags behind contrastive decoding methods in COVER score for LLaVA, it ranks second for the InstructBLIP model. This slight reduction in coverage is accompanied by a substantial decrease in hallucination, representing a favorable trade-off in many real-world applications. Notably, even with LLaVA, CAAC still outperforms OPERA—the strongest baseline in terms of hallucination metrics.

\paragraph{POPE.} 
The Polling-based Object Probing Evaluation (POPE) benchmark \citep{li_evaluating_2023} provides a streamlined approach to assess object hallucination in Large Vision-Language Models by querying whether specific objects exist in a given image. POPE employs three sampling settings for negative samples: random, popular, and adversarial, each designed to challenge the model’s discriminative capabilities differently. Although our CAAC framework is primarily designed for generative tasks, it exhibits robust performance in this discriminative setting, as shown in \cref{tab:pope_results}. CAAC achieves Accuracy and F1 scores within 1\% of OPERA and outperforming all other baselines. These results highlight CAAC’s effectiveness in mitigating hallucinations beyond its generative focus.
% , outperforming or matching baseline methods, thus demonstrating its versatility and robustness.

% \begin{figure}[t]
%   \centering
%   \begin{subfigure}{0.23\textwidth}
%     \includegraphics[width=\linewidth]{figs/chair_ablation_barplot.png}
%     \caption{}
%     \label{fig:ablation_chair}
%   \end{subfigure}
%   \hfill
%   \begin{subfigure}{0.23\textwidth}
%     \includegraphics[width=\linewidth]{figs/amber_ablation_barplot.png}
%     \caption{}
%     \label{fig:ablation_amber}
%   \end{subfigure}
%   \caption{Ablation study results for InstructBLIP on the (a) CHAIR and (b) AMBER benchmarks. The plots show the performance of the baseline, VTC-only, AAR-only, and full CAAC (VTC + AAR) settings in terms of hallucination rates and recall metrics.}
%   \label{fig:ablation}
% \end{figure}

\begin{table}[t]\small
  \centering
  % % \vspace{-10pt}
  \caption{Ablation Study Summary for InstructBLIP on CHAIR and AMBER Benchmarks}
  \begin{tabular}{lcc|ccc}
    \toprule
    \multirow{2}{*}{Configuration} & \multicolumn{2}{c|}{CHAIR} & \multicolumn{3}{c}{AMBER} \\
    \cmidrule{2-6}
    & $\mathrm{C}_i$$\downarrow$ & $\mathrm{C}_s$$\downarrow$  & CHAIR$\downarrow$ & HAL$\downarrow$ & COVER$\uparrow$ \\
    \midrule
    Base Model & 16.6 & 55.6  & 9.6 & 36 & 46.5 \\
    VTC-only & 11.3 & 40.2  & 6.1 & 27.6 & 48.3 \\
    AAR-only & 13.4 & 50.2  & 7.8 & 36.2 & \textbf{52.5} \\
    VTC+AAR & \textbf{10.8} & \textbf{37.4}  & \textbf{5.6} & \textbf{25.8} & 47.8 \\
    \bottomrule
  \end{tabular}
  % % \vspace{-8pt}
  % \caption*{Full results and visualizations are in the supplementary material (Sec. 5).}
  \label{tab:ablation_summary}
\end{table}

\subsection{Ablation Study}

To measure the influence of each module within the CAAC framework, we conducted ablation experiments using the InstructBLIP model on the AMBER and CHAIR benchmarks. We evaluated the base model, VTC-only, AAR-only, and the full CAAC framework with both modules. The results on CHAIR and AMBER benchmarks are presented in \cref{tab:ablation_summary}. As shown, both modules individually contribute to lowering hallucination rates, as measured by CHAIR and Hal metrics, while also increasing coverage and recall compared to the base model. Also, the full CAAC framework achieves the most significant improvements overall.

\subsection{Hyperparameter Analysis}
We optimized the CAAC framework by tuning its key parameters, focusing on the Adaptive Attention Re-Scaling (AAR) and Visual-Token Calibration (VTC) modules to balance hallucination reduction while preserving response quality and integrity. For AAR, we set the confidence threshold \( p_{\text{thr}} = 0.25 \), \( \lambda_{\max} = 1.5 \), and applied it to all decoding layers, achieving consistent and coherent outputs. For VTC, applying it to the first 10 layers (out of 32) minimized hallucination rates effectively, avoiding the incoherence or truncated sequences observed with full-layer application. The smoothing parameter \( \beta \) was found to be very impactful. Large values of \( \beta \) ($ \geq 0.9)$ often resulted in impaired generation sequences. However, intermediate values for \( \beta \), \( 0.3 \sim 0.7\), resulted in coherent and high-quality responses. A more comprehensive analysis of the models' settings is provided in the technical appendix (Sec. 4).

\section{Discussion}

\textbf{Inference Efficiency.} CAAC’s dual-pass mechanism introduces modest latency, justified by improved factuality. This overhead is common among hallucination mitigation methods; contrastive decoding (e.g., VCD, M3ID, AvisC) requires two passes \emph{per token}, while CAAC triggers a second pass for only ~14\% of tokens on 500 MS COCO images. A detailed runtime comparison is in the technical appendix (Sec. 1), showing CAAC’s competitive efficiency.
\textbf{Faithfulness vs. Completeness.} CAAC aims to improve the faithfulness of LVLM outputs, which naturally introduces a trade-off with completeness -- a trend observed across nearly all hallucination mitigation methods. For example, OPERA, the best-performing baseline in hallucination metrics (CHAIR and HAL), ranks lowest in average COVER score (Table~\ref{tab:amber_results}). In contrast, methods like VCD, M3ID, and AvisC attain higher coverage but perform worse on hallucination reduction. CAAC, however, strikes a strong balance: it substantially reduces hallucinations across all benchmarks while consistently improving upon the base model in COVER and scoring second on InstructBLIP. This demonstrates CAAC's ability to suppress hallucinations while preserving a reasonably high level of completeness. Moreover, this trade-off is acceptable in many real-world applications such as medical report generation~\cite{hartsock_vision-language_2024}, legal document analysis~\cite{magesh_hallucination-free_2025}, and autonomous driving~\cite{keskar_evaluating_nodate}, where factual consistency takes precedence over exhaustive content.

% \paragraph{Generalization and Consistency.}
\textbf{Generalization and Consistency.} CAAC is particularly effective in long-form generation tasks, where hallucinations tend to arise in later tokens (\cref{fig:rel-vs-pos}). The AAR module is designed to intervene selectively during these later stages -- when the model’s attention begins to drift away from image tokens -- making it well-suited for scenarios where uninterrupted generation is required. CAAC consistently outperforms strong baselines on generative benchmarks such as CHAIR and AMBER (Tables~\ref{tab:chair_results},~\ref{tab:amber_results}) by a notable margin in hallucination metrics, while remaining within 1\% of the top-performing baseline on discriminative benchmarks like POPE. This balance highlights the effectiveness of CAAC’s design. Furthermore, improvements remained consistent on a more advanced LVLM like LLaVA-NeXT (\cref{tab:amber_results}), demonstrating CAAC’s adaptability across architectures.

\section{Conclusion}
We introduced the Confidence-Aware Attention Calibration (CAAC), a training-free framework that mitigates hallucination in LVLMs by addressing spatial and modality biases through Visual-Token Calibration and Adaptive Attention Re-Scaling, ensuring consistent visual grounding across diverse generation tasks. Experiments on benchmarks like CHAIR, AMBER, and POPE MSCOCO demonstrate CAAC’s effectiveness in reducing hallucination rates, surpassing baselines like OPERA, particularly in long sequences, despite a trade-off with metrics like COVER and Recall. This prioritization of factual accuracy over exhaustive detail makes CAAC a practical solution for enhancing LVLM reliability in safety-critical applications.

\bibliography{main}


% Ensure this is included in the preamble of your main document
% If not already present, add:
% \usepackage{xcolor}
% \usepackage{aaai2026} % Already in your main document
% \usepackage{times,helvet,courier} % If not already included

% Checklist definitions
\makeatletter
\newif\ifreproStandalone
\reproStandalonefalse % We're including this in the main paper
\makeatother

\setlength{\leftmargini}{20pt}
\makeatletter
\def\@listi{\leftmargin\leftmargini \topsep .5em \parsep .5em \itemsep .5em}
\def\@listii{\leftmargin\leftmarginii \labelwidth\leftmarginii \advance\labelwidth-\labelsep \topsep .4em \parsep .4em \itemsep .4em}
\def\@listiii{\leftmargin\leftmarginiii \labelwidth\leftmarginiii \advance\labelwidth-\labelsep \topsep .4em \parsep .4em \itemsep .4em}
\makeatother

\setcounter{secnumdepth}{0}
\renewcommand\thesubsection{\arabic{subsection}}
\renewcommand\labelenumi{\thesubsection.\arabic{enumi}}

\newcounter{checksubsection}
\newcounter{checkitem}[checksubsection]

\newcommand{\checksubsection}[1]{%
  \refstepcounter{checksubsection}%
  \paragraph{\arabic{checksubsection}. #1}%
  \setcounter{checkitem}{0}%
}

\newcommand{\checkitem}{%
  \refstepcounter{checkitem}%
  \item[\arabic{checksubsection}.\arabic{checkitem}.]%
}
\newcommand{\question}[2]{\normalcolor\checkitem #1 #2 \color{blue}}
\newcommand{\ifyespoints}[1]{\makebox[0pt][l]{\hspace{-15pt}\normalcolor #1}}

\newpage

\end{document}